\pdfoutput=1

\documentclass[11pt]{article}

\usepackage[final]{acl}

\usepackage{times}
\usepackage{latexsym}
\usepackage{amsmath}

\usepackage[T1]{fontenc}

\usepackage[utf8]{inputenc}

\usepackage[nopatch=footnote]{microtype}

\usepackage{inconsolata}
\usepackage{graphicx}

\usepackage{xcolor}

\usepackage{philex}
\usepackage{multirow}
\usepackage{adjustbox}
\usepackage{booktabs}
\usepackage[nameinlink]{cleveref}
\usepackage{float}
\usepackage{tabularx}
\usepackage{svg}
\usepackage{xcolor}
\usepackage{dsfont}

\crefname{section}{\S}{\S\S}
\Crefname{section}{\S}{\S\S}
\restylefloat{table}

\title{Rolling the DICE on Idiomaticity: How LLMs Fail to Grasp Context}
\author{ 
{ \bf {Maggie Mi}}$^1$ $\-$ $\-$ $\-$ {\bf {Aline Villavicencio}}$^{1,2,3}$ $\-$ $\-$ $\-$ {\bf {Nafise Sadat Moosavi}}$^1$ \\
{$^{1}${University of Sheffield} $\-$ $\-$ $\-$ $\-$ $^2${University of Exeter} $\-$ $\-$ $\-$ $\-$ $^3${Alan Turing Institute}}\\ 
\texttt{{zmi1@sheffield.ac.uk}}$\-$ $\-$ $\-$ $\-$  \texttt{{a.villavicencio@exeter.ac.uk}}$\-$ $\-$ $\-$ $\-$  \\
 \texttt{{n.s.moosavi@sheffield.ac.uk}}
}

\newcommand{\squishlist}{
\begin{list}{$\bullet$}
{   \setlength{\itemsep}{0pt}
   \setlength{\parsep}{3pt}
   \setlength{\topsep}{3pt}
   \setlength{\partopsep}{0pt}
   \setlength{\leftmargin}{1.5em}
   \setlength{\labelwidth}{1em}
   \setlength{\labelsep}{0.5em} } }
\newcounter{Lcount}
\newcommand{\squishlisttwo}{
\begin{list}{\arabic{Lcount}. }
  { \usecounter{Lcount}
 \setlength{\itemsep}{0pt}
 \setlength{\parsep}{0pt}
 \setlength{\topsep}{0pt}
 \setlength{\partopsep}{0pt}
 \setlength{\leftmargin}{2em}
 \setlength{\labelwidth}{1.5em}
 \setlength{\labelsep}{0.5em} } }
\newcommand{\squishend}{\end{list} }

\begin{document}
\maketitle
\begin{abstract}

Human processing of idioms heavily depends on interpreting the surrounding context in which they appear. While large language models (LLMs) have achieved impressive performance on idiomaticity detection benchmarks, this success may be driven by reasoning shortcuts present in existing datasets. To address this, we introduce a novel, controlled contrastive dataset (DICE) specifically designed to assess whether LLMs can effectively leverage context to disambiguate idiomatic meanings. Furthermore, we investigate the influence of collocational frequency and sentence probability—proxies for human processing known to affect idiom resolution—on model performance. Our results show that LLMs frequently fail to resolve idiomaticity when it depends on contextual understanding, and they perform better on sentences deemed more likely by the model. Additionally, idiom frequency influences performance but does not guarantee accurate interpretation. Our findings emphasize the limitations of current models in grasping contextual meaning and highlight the need for more context-sensitive evaluation. 

\vspace{.75pt}

\includegraphics[width=1.25em,height=1.25em]{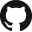}\hspace{0.75em}%
{\footnotesize\url{https://github.com/mi-m1/dice}}

\end{abstract}

\section{Introduction}
\label{section:introduction}
Idiomatic expressions (IEs) are strange birds whose meaning may not be straightforwardly related to the meaning of the component words in isolation. For example, proficient English speakers understand ``\textit{kick the bucket}'' not as ``\textit{striking a metal container}'', but as ``\textit{to die}''. Estimates suggest that there are 25,000 fixed expressions in English alone \citep{Weinreich1969}, and a similar estimate is quoted for French \citep{gross1982}. Notably, this figure is comparable to the order of magnitude of individual words in the lexicon \citep{jackendoff1997architecture}. This suggests that idioms are not mere linguistic curiosities but fundamental components of language. 

Additionally, some of these expressions are ambiguous, ``potentially idiomatic expressions" (PIEs) that can be interpreted either non-compositionally (figuratively or idiomatically) or compositionally (literally), depending on the context in which they appear. Accurately identifying the meaning of a PIE often depends on its context and is essential for numerous downstream applications, such as machine translation \citep{dankers-etal-2022-transformer, barreiro-etal-2013-multiwords, salton-etal-2014-empirical, fadaee-etal-2018-examining}, sentiment analysis \citep{WILLIAMS20157375, liu-etal-2017-idiom}, and automatic spelling correction \cite{horbach-etal-2016-corpus}. Beyond these applications, understanding idiomatic expressions in context is also crucial to grasp the underlying meaning of the text.

Existing datasets that feature expressions with both literal and idiomatic usages often fail to rigorously examine the role of context in disambiguating meaning. This shortcoming arises because the literal meanings in such datasets typically result from syntactic modifications or semantic shifts, which inherently disrupt the idiomaticity of the expression. In fact, these changes in form have been deliberately used as signals to disentangle and differentiate literal from figurative meaning \cite{fazly-etal-2009-unsupervised}. For example, passivisation (2) or modification to the expression (3) often strip the expression of its idiomatic meaning, as shown in comparison to the idiomatic usage in (1) \cite{jackendoff1997architecture,gibbs-synatic-frozeness,langlotz-2006,kyriacou-2020}\footnote{``\textit{Kick the bucket}'' is a classic example of a rigid idiom, but expressions vary in flexibility; some can better tolerate changes without losing their idiomatic meaning than others.}. 
\lb{1}{He \textbf{kicked the bucket.}}
\lb{2}{The \textbf{bucket} \textcolor{blue}{was} \textbf{kicked} \textcolor{blue}{by him}.}
\lb{3}{He \textbf{kicked} the \textcolor{blue}{tin} \textbf{bucket} \textcolor{blue}{angrily}.}

Consequently, this allows models to exploit surface-level differences, such as changes in grammatical structure or insertions, as shortcuts for determining literalness, rather than relying on their understanding of idiomaticity in context as the task intends. This undermines the evaluation's effectiveness by encouraging models to use superficial cues instead of deeper contextual comprehension.

In this paper, we address this gap by introducing \emph{DICE} (Dataset for Idiomatic Contrastive Evaluation), a benchmark designed to evaluate the ability of LLMs to interpret idiomatic expressions in context. We focus on idioms, as these figurative expressions may serve as key indicators of a model's linguistic understanding. Given the possibility of a predominant sense, DICE presents idiomatic expressions in both literal and figurative contexts and challenges models to rely on context for the correct interpretation. This contrastive evaluation forces models to distinguish between meanings based solely on context, preventing them from relying on memorized idioms. We hypothesize that if models do not depend on memorization, they will perform equally well in both senses of a PIE.


Moreover, in human language processing, factors such as frequency and familiarity \citep{swinney1979access,nippold1993familiarity}, alongside context \citep{estill1982interpreting,gibbs1989kick,gibbs-idiom-decomposition}, are known to influence idiom comprehension and processing speed \citep{tabossi2009idioms}. We investigate whether similar effects hold for language models. Specifically, beyond contextual understanding, we examine the influence of a language\-intrinsic feature (expression frequency) and a model\-intrinsic feature (sentence likelihood) on model performance. 

\paragraph{Contributions} 
(1) We present DICE, a comprehensive and robust evaluation dataset, containing PIEs that occur in the same grammatical form across both figurative and literal contexts.
(2) Through fine-grained evaluations, we find that models struggle to use context for idiomaticity processing.
(3) Based on frequency estimates, we find that frequency is not a ``free lunch'': whilst highly frequent idioms may be more likely to be disambiguated correctly, there is a trade-off in model performance between literal and figurative settings.
(4) For models that demonstrate some capacity for idiomatic understanding, we observe a strong relationship between the likelihood of the contextual sentence and idiomaticity detection performance.

\section{Related Works}
\label{sec:related}

\paragraph{Idiomaticity Detection.} The task of idiomaticity sense disambiguation (ISD), or idiomaticity detection, involves evaluating whether an expression is used literally or figuratively in a sentence \citep{liu-hwa-2018-heuristically, salehi-etal-2014-detecting, senaldi-etal-2016-lexical,gharbieh-etal-2016-word}. This task is typically framed as binary classification. While large language models have achieved strong performance on existing ISD benchmarks \citep{phelps-etal-2024-sign, zengSemanticCompatability}, it remains unclear whether this reflects true contextual understanding or reliance on memorized surface forms \citep{garcia-etal-2021-probing}. Given the crucial role of context in resolving idiomaticity, it is essential to evaluate whether models are genuinely interpreting surrounding text or simply exploiting distributional cues.

\paragraph{Existing Datasets.} The biggest dataset for idiomatic sense disambiguation is MAGPIE \citep{haagsma-etal-2020-magpie}. MAGPIE contains a total of 56,622 PIE instances, across 1,756 idioms. However, a large amount of deviation of the form of an expression was allowed when MAGPIE was curated. As a result, most of the literal uses of PIEs involve modifications to the form of the expression (similar to the example in \Cref{section:introduction}). Other large datasets targeting various types of IEs have been released: The VNC-Tokens dataset focusing on verb-noun combinations~\citep{cook2008vnc},  IDIX on verb-noun phrase or verb-prepositional phrase expressions \citep{sporleder-etal-2010-idioms}. SemEval-2013 has unrestricted expressions \citep{korkontzelos-etal-2013-semeval}, AStitchInLanguageModels contains noun compounds \citep{tayyar-madabushi-etal-2021-astitchinlanguagemodels-dataset} and more recently, IdioTS contains a mixture of expressions changed and unchanged to support the literal meaning \citep{de-luca-fornaciari-etal-2024-hard}.

To address these problems: (1) we propose a novel evaluation set (DICE), which strictly controls the form of idiomatic expressions. This design eliminates the possibility that models rely on grammatical variations for idiomaticity disambiguation. By maintaining consistent forms across literal and figurative  contexts, DICE ensures that the challenge lies in understanding contextual nuances, thereby providing a more accurate assessment of a model's idiomatic comprehension. (2) Existing datasets typically focus on a single type of expression. We address this limitation by including both phrasal idioms and noun compounds to provide broader coverage of idiomatic phenomena.

\paragraph{Contrastive Evaluation.}
This evaluation paradigm involves comparing model outputs on carefully constructed input pairs that differ in minimal, controlled ways, typically along a dimension relevant to a specific linguistic phenomenon \citep{prasad-etal-2019-using,garcia-etal-2021-assessing}. Such comparisons have been found to be particularly effective at isolating specific linguistic competencies or revealing systematic weaknesses in generalization and robustness \citep{10.1162/tacl_a_00115, sennrich-2017-grammatical, robertson-2019-contrastive}.
Our work adopts this approach to assess contextual comprehension in idiomaticity detection. 
Specifically, DICE presents potentially idiomatic expressions in both literal and figurative uses while holding surface form constant. This setup requires models to rely entirely on surrounding context to disambiguate meaning, eliminating the possibility of exploiting shallow lexical or syntactic cues. 

\paragraph{Memorization and Context.} 
Pretrained language models tend to handle IEs mainly through memorization of stored expressions and token distributions rather than reasoning about the meaning in context. This reliance on memorization becomes evident when models face novel noun compounds. For instance, \citet{li-etal-2022-systematicity} found that while GPT-3’s interpretations of novel compounds closely matched human responses, its performance faltered when tasked with interpreting nonsensical strings, which suggests limited contextual flexibility.

Similarly, \citet{coil-shwartz-2023-chocolate} explored noun compound interpretation using GPT-3 and found that, although the model performs well with common noun compounds, its performance drops significantly with novel compounds. This suggests that the model relies on pre-existing knowledge for familiar compounds, but struggles when required to reason about unseen combinations. 
Supporting this view, \citet{cheng-bhat-2024-context} found that models sometimes perform better on idiomatic reasoning tasks when contextual information is removed, implying that context may occasionally mislead rather than assist the model. \citet{sun-etal-2021-long} further showed that LLMs tend to use context only when the relevant information is explicitly present. These findings collectively raise questions about whether current models truly engage in contextual reasoning or primarily depend on surface-level cues.

\begin{figure}[!htb]
  \includegraphics[width=\columnwidth]{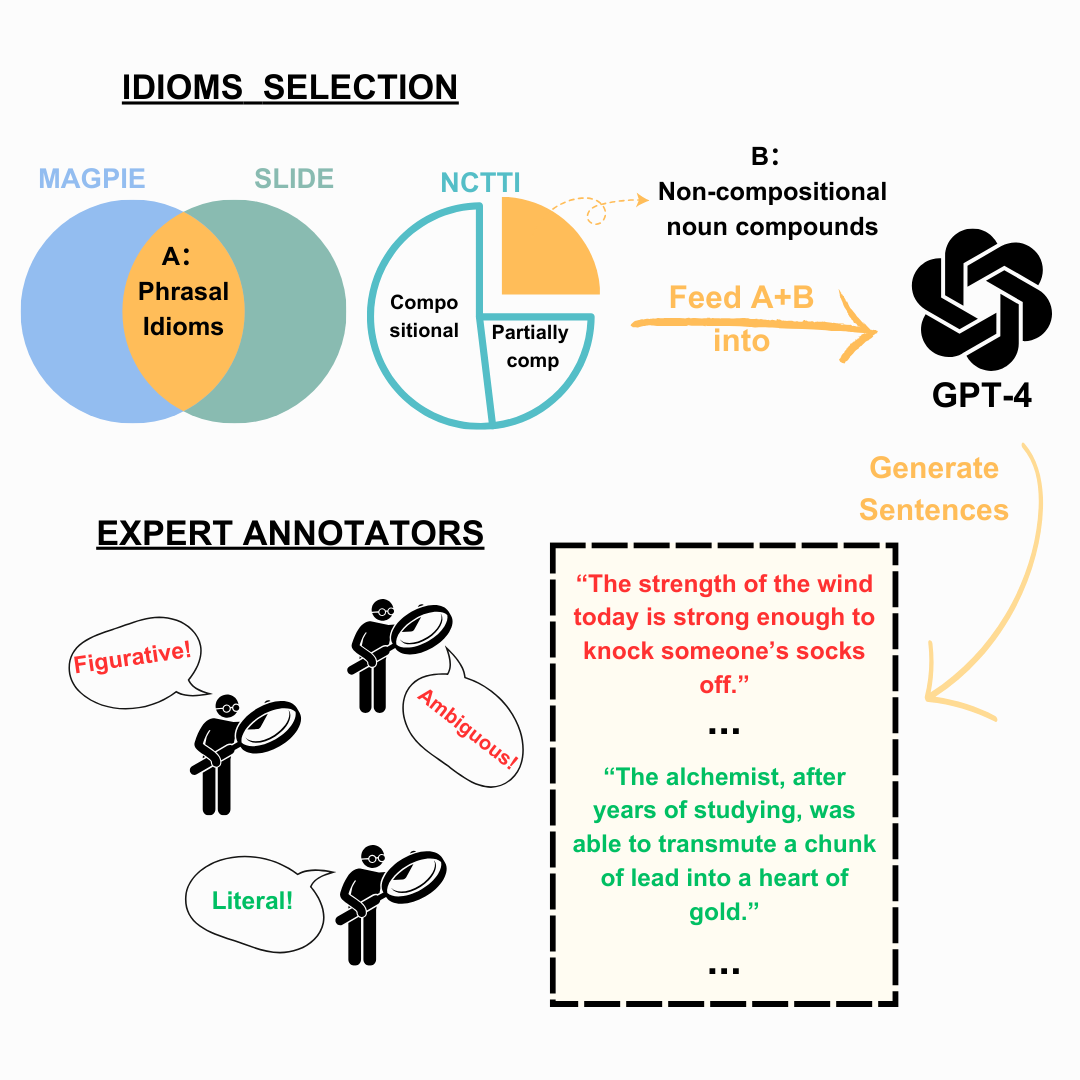}
  \vspace{-2mm}
  \caption{Overview of the DICE curation process. A list of idioms is extracted from existing datasets, and GPT-4 is prompted to generate sentences using each idiom in a literal context. Experts annotate these to form the literal subset. Figurative examples are drawn directly from existing datasets.}
  \label{fig:data_curation}
\end{figure}

\section{DICE: Dataset for Idiomatic Contrastive Evaluation }
\label{sec:DICE}

We now describe the construction of DICE. \Cref{fig:data_curation} shows an overview of the curation process.

\begin{table*}[!htb]
    \centering
    \adjustbox{max width=\textwidth}{
        \begin{tabular}{lll}
        \toprule
        &Counts &Examples and Remarks \\\midrule
        Number of Sentences (Literal) &1033& Carpenters recommend not to sand against the grain as it can damage the wood.\\
        Number of Sentences (Figurative) &1033 & Out of duty she had caved in, but it still went against the grain. (MAGPIE) \\
        Total no. of sentences &2066 &- \\
        Number of Unique Idioms &402 &- \\
        Total Number of Expressions &402 &103 noun compounds + 299 phrasal expressions \\
        Average length of sentences (literal) &15.4 words&- \\
        Average length of sentences (figurative) &28.1 words &- \\
        \bottomrule
        \end{tabular}}
    \vspace{-2mm}
    \caption{Summary statistics of the DICE dataset.}
    \label{table:summary_stats}
\end{table*}

\paragraph{Expression Selection.} To build our dataset, we compiled a comprehensive list of idiomatic expressions, focusing on both phrasal idioms and noun compound idioms. For phrasal idioms, we identified overlapping expressions from two established resources: MAGPIE \citep{haagsma-etal-2020-magpie} and SLIDE \citep{jochim-etal-2018-slide}. For noun compound idioms, we selected idiomatic expressions that appeared in both the NCTTI dataset \citep{garcia-etal-2021-probing} and AStitchInLanguageModels \citep{tayyar-madabushi-etal-2021-astitchinlanguagemodels-dataset}. We specifically focused on idiomatic noun compounds, as the meaning of these expressions cannot be directly derived from the meanings of their individual words.

We excluded compositional and partially compositional compounds, as these tend to have dominant literal meanings that are difficult to override in context (e.g., ``\textit{skin tone}'' or ``\textit{noble gas}''). Such expressions do not present the same interpretive challenge for models, as their meanings are more directly tied to their components. By focusing solely on non-compositional idioms, we ensure that the dataset tests the model's ability to interpret figurative language based on context rather than relying on straightforward lexical composition. This selection process resulted in a total of 783 unique idiomatic expressions, comprising 680 phrasal idioms and 103 idiomatic noun compounds.

\paragraph{Sentence Generation.} We used GPT-4 \citep{openai2024gpt4} to generate sentences where a given idiom appears in a literal context, intentionally suppressing its figurative interpretation. The specific prompting setup we used for sentence generation is provided in \Cref{appendix:appendix_prompt_gen}. To determine the best model for sentence generation, we piloted the process using three versions: GPT-4o, GPT-4, and GPT-3.5. The models were prompted to produce three different sentences, where the form of the idiom must be kept the same.
Overall, we found GPT-4 to perform the best in generating sentences where the figurative interpretation is suppressed. Our preference for GPT-4 aligns with the findings of \citet{phelps-etal-2024-sign}, which demonstrate that off-the-shelf GPT-4 possesses relatively stronger idiomaticity knowledge as it performed consistently well across idiomaticity detection tasks compared to other off-the-shelf LLMs. In total, we obtained 2,349 sentences from GPT-4.

\paragraph{Expert Annotations.}
 We recruited four experts with at least three years of university-level experience in Linguistics, compensated at a rate of £15/hour. The annotators reviewed each sentence and decided whether to accept it unconditionally, skip it, or reject it if the idiom’s figurative meaning could not be fully suppressed. In cases of rejection, annotators provided reasons such as ambiguity, figurative interpretation, change of form, or other issues. We provide information about this subset in \Cref{appendix:addtional_summary_stats} along with the briefing given to annotators in \Cref{appendix:briefing}. If an expression was skipped, a second annotator reviewed it to confirm if it should be discarded. Examples of sentences for each category are presented in \Cref{table:annotations}. The inter-rater agreement was high, as indicated by a Cohen's kappa coefficient of 0.95.

The figurative counterparts of these sentences were sourced from MAGPIE and AStitchInLanguageModels. We ensure that the same number of variants is matched between the figurative and literal settings. In other words, if we have three sentences containing ``all hell broke loose'' in literal contexts, we would extract an equal number of sentences containing the idiom from the figurative datasets. This ensures that the dataset is balanced with regards to the idiomatic and literal interpretation  of each expression.
 
\subsection{Dataset Statistics}
In total, DICE consists of 2,066 sentences, featuring 402 expressions. A summary of its statistics is presented in \Cref{table:summary_stats}. In this paper, we only use the subset whose literal meaning was confirmed by the annotators. However, we release the complete dataset, along with an additional subset that can serve as a  resource for additional analyses or for creating even more challenging evaluation settings (see \Cref{appendix:addtional_summary_stats}).

\section{How Well Do LLMs Use Context for Idiomaticity?}
\label{sec:eval}
Using DICE, we evaluated the ability of various language models to differentiate between literal and figurative uses of idioms. Replicating the Idiomaticity Sense Disambiguation (ISD) task, we prompted each model with a sentence containing an idiom and asked it to classify the idiom as either 
``literal'' or ``figurative'' based on its usage in context.

\subsection{Experimental Setup.}
\label{ref:experimental_setup}
\paragraph{Models.} We evalute 13 models on the task of idiomatic sense disambiguation. These models are: GPT-4o, GPT-3.5-Turbo \citep{brown2020gpt3}, FLAN-T5 models in the XXL (11B), XL (3B), Large (780M), Small (80M) sizes \citep{chung-etal-2023-increasing}, Llama 2 (7B, 13B, 70B) \citep{touvron2023llama}, and Llama 3 models (incl. Llama 3.1 (405B, 8B, 70B) \citep{dubey2024llama3herdmodels}. Additionally, we evaluated GPT-4 \citep{openai2024gpt4}, which was used to generate the sentences. The computational resources used are reported in \Cref{sec:implementation_details}. 

\paragraph{Prompts.} 

We used three prompt variations (\Cref{tab:eval_prompts} in \Cref{sec:prompting_paradigm}) that are semantically equivalent but differ in surface form. This allows us to evaluate the robustness of model predictions to prompt phrasing. We report the mean and standard deviation across these prompts to assess performance stability under variation. We also run experiments in a few-shot setting, where an annotated example (shown in the middle part of \Cref{tab:eval_prompts}) is prepended to each prompt.

\paragraph{Evaluation.}
To thoroughly evaluate the models' performance, we employed three distinct evaluation settings. 

\textbf{Accuracy} includes two sub evaluations:
\squishlist
\item \textbf{Figurative Accuracy}: We compute the accuracy of each model in correctly identifying the figurative uses of expressions within the figurative subset. Let $F$ be the set of all figurative instances, $y_i$ and $\hat{y}_i$ denote the true and predicted labels, respectively.

\begin{equation}
\text{Accuracy}_{\text{fig}} = \frac{1}{|F|} \sum_{i \in F} \mathds{1}(\hat{y}_i = y_i)
\end{equation}

where
\[
\mathds{1}(\text{condition}) =
\begin{cases}
1 & \text{if condition is true} \\
0 & \text{otherwise}
\end{cases}
\]
\item \textbf{Literal Accuracy}: We assessed the accuracy of the models in correctly identifying the literal uses of expressions within the literal subset. These evaluations measure the models' ability to recognize idiomatic and literal meanings based on context. Let $L$ be the set of all literal instances.
\squishend

\begin{equation}
\text{Accuracy}_{\text{lit}} = \frac{1}{|L|} \sum_{i \in L} \mathds{1}(\hat{y}_i = y_i)
\end{equation}

\begin{table*}
    \centering
    \small
    \scalebox{0.85}{
        \begin{tabular}{l|rrrrrrr}
            \toprule
             \multirow{2}{*}{\textbf{Model}} & \multicolumn{3}{c}{\textbf{Accuracy}} & \multicolumn{3}{c}{\textbf{Lenient Consistency}} & \multicolumn{1}{c}{\textbf{Strict Consistency}} \\
             \cmidrule(lr){2-4} \cmidrule(lr){5-7} \cmidrule(lr){8-8} \\ 
             & Figurative & Literal & Overall & Figurative & Literal & Overall & Both Settings \\
             \midrule
            \multicolumn{8}{c}{\textbf{\textsc{Zero-shot}}} \\\midrule
            GPT-4o &87.05 ± 3.62 &87.30 ± 2.98 &84.33 ± 4.44 &69.49 ± 11.71 &71.06 ± 6.68 &70.32 ± 7.11 &48.59 ± 9.75 \\
            GPT-3.5 Turbo &79.05 ± 5.01 &70.02 ± 12.72 &75.54 ± 7.81 &82.59 ± 9.17 &44.36 ± 22.28 &63.47 ± 7.61 &32.84 ± 15.81 \\
            Flan-T5-XXL (11B) &77.18 ± 1.40 &74.91 ± 8.35 &76.40 ± 4.49 &63.93 ± 13.71 &58.79 ± 23.16 &61.36 ± 4.73 &32.92 ± 6.80 \\
            Flan-T5-XL (3B) &70.48 ± 3.56 &33.94 ± 26.91 &59.65 ± 8.19 &91.13 ± 6.97 &13.02 ± 11.24 &52.07 ± 3.58 &9.95 ± 8.88 \\
            Flan-T5-Large (780M) &66.63 ± 0.10 &3.45 ± 4.72 &50.42 ± 0.53 &97.68 ± 3.40 &0.58 ± 0.80 &49.13 ± 1.30 &0.58 ± 0.80 \\
            Flan-T5-Small (80M) &0.51 ± 0.59 &66.72 ± 0.07 &50.13 ± 0.15 &0.00 ± 0.00 &100.00 ± 0.00 &50.00 ± 0.00 &0.00 ± 0.00 \\
            Llama 3.1 (405B) &88.63 ± 2.36 &88.25 ± 3.93 &\textbf{88.45 ± 3.10} &78.52 ± 5.61 &80.02 ± 12.43 &\textbf{79.27 ± 3.46} & \bfseries 60.36 ± 6.61 \\
            Llama 3 (70B) &87.72 ± 4.63 &86.13 ± 7.10 &87.00 ± 5.73 &81.84 ± 4.00 &72.64 ± 16.12 &77.24 ± 7.45 &57.55 ± 12.41 \\
            Llama 3 (8B) &79.27 ± 1.97 &74.01 ± 2.79 &76.91 ± 2.25 &77.86 ± 5.18 &48.76 ± 3.37 &63.31 ± 1.43 &33.83 ± 2.60 \\
            Llama 2 (70B) &76.28 ± 4.39 &56.64 ± 17.13 &69.62 ± 7.82 &93.20 ± 4.75 &24.54 ± 16.89 &59.12 ± 5.78 &21.81 ± 13.51 \\
            Llama 2 (13B) &68.99 ± 1.39 &36.09 ± 3.85 &58.26 ± 1.96 &85.41 ± 3.56 &8.37 ± 3.34 &46.93 ± 2.30 &5.64 ± 2.00 \\
            Llama 2 (7B) &55.51 ± 19.54 &31.97 ± 24.25 &51.34 ± 1.55 &59.87 ± 46.26 &18.08 ± 29.16 &38.97 ± 8.59 &1.66 ± 1.37 \\
            \midrule
            GPT-4 &88.56 ± 2.03 &88.63 ± 2.08 &88.48 ± 2.18 &79.02 ± 3.11 &78.03 ± 4.60 &78.52 ± 2.95 &59.62 ± 4.67 \\
            \midrule
                    \multicolumn{8}{c}{\textbf{\textsc{One-shot}}} \\\midrule
            GPT-4o &89.43 ± 1.23 &90.15 ± 1.71 &\textbf{89.72 ± 1.45} &74.63 ± 1.99 &87.40 ± 5.81 &\textbf{81.01 ± 1.93} & \textbf{63.52 ± 3.15} \\
            GPT-3.5 Turbo &79.41 ± 4.19 &72.69 ± 10.87 &76.70 ± 6.54 &78.44 ± 8.80 &49.42 ± 18.96 &63.93 ± 5.92 &34.16 ± 12.19 \\
            Flan-T5-XXL (11B) &10.20 ± 15.69 &67.90 ± 1.91 &52.79 ± 4.34 &1.58 ± 2.52 &99.25 ± 1.29 &50.41 ± 0.61 &1.49 ± 2.37 \\
            Flan-T5-XL (3B) &0.64 ± 0.80 &66.71 ± 0.11 &50.13 ± 0.22 &0.08 ± 0.14 &99.83 ± 0.29 &49.96 ± 0.19 &0.08 ± 0.14 \\
            Flan-T5-Large (780M) &3.28 ± 3.64 &66.27 ± 0.45 &50.00 ± 0.00 &0.66 ± 0.76 &96.93 ± 3.73 &48.80 ± 1.48 &0.00 ± 0.00 \\
            Flan-T5-Small (80M) &45.23 ± 39.19 &35.55 ± 33.55 &53.03 ± 5.25 &60.78 ± 53.37 &37.31 ± 54.62 &49.05 ± 1.65 &2.40 ± 4.16 \\
            Llama 3.1 (405B) &89.57 ± 1.80 &89.54 ± 2.54 &89.53 ± 2.17 &79.10 ± 3.26 &82.01 ± 7.85 &80.56 ± 2.56 &63.27 ± 4.66 \\
            Llama 3 (70B) &87.75 ± 3.76 &86.97 ± 5.64 &87.27 ± 4.61 &78.52 ± 3.59 &75.62 ± 14.01 &77.07 ± 6.00 &57.55 ± 10.22 \\
            Llama 3 (8B) &80.32 ± 5.33 &73.81 ± 11.40 &77.59 ± 7.62 &79.35 ± 1.08 &48.01 ± 15.70 &63.68 ± 7.34 &34.91 ± 13.59 \\
            Llama 2 (70B) &70.40 ± 1.19 &31.44 ± 6.18 &58.65 ± 2.28 &96.52 ± 0.66 &7.55 ± 2.75 &52.03 ± 1.50 &6.72 ± 2.63 \\
            Llama 2 (13B) &70.64 ± 1.15 &34.20 ± 6.92 &59.36 ± 2.45 &94.94 ± 0.52 &9.54 ± 4.14 &52.24 ± 1.83 &8.29 ± 3.11 \\
            Llama 2 (7B) &70.26 ± 3.14 &42.18 ± 26.31 &61.21 ± 9.28 &80.76 ± 15.43 &20.73 ± 22.25 &50.75 ± 3.46 &11.69 ± 10.42 \\
            \midrule
            GPT-4 &88.52 ± 1.49 &88.95 ± 2.09 &88.42 ± 1.73 &78.44 ± 0.76 &77.94 ± 5.84 &78.19 ± 2.63 &58.87 ± 4.86 \\
            \bottomrule
        \end{tabular}
        }
    \caption{Results are reported as mean $\pm$ standard deviation over 3 different prompt variants, under zero-shot (top) and one-shot (bottom) conditions.  GPT-4 is shown separately, as it generated the evaluation sentences.}
    \label{table:main_results_table}
\end{table*}

\textbf{Lenient Consistency}: This metric rewards the model for consistently classifying all instances of an expression as either literal or figurative within a specific type. For example, if the model correctly classifies all literal variations of a given expression as ``literal'', it earns a point for that expression. Similarly, the model is rewarded if it correctly identifies all figurative instances of an expression as ``idiomatic''.
Let $E$ be the set of all expressions. Let $L_e$ and $F_e$ be the literal and figurative instances of expression $e$, respectively.

\begin{equation}
C_e^{\text{lit}} =
\begin{cases}
1 & \text{if } \forall i \in L_e, \hat{y}_i = y_i  \\
0 & \text{otherwise}
\end{cases}
\end{equation}

\begin{equation}
C_e^{\text{fig}} =
\begin{cases}
1 & \text{if } \forall i \in F_e, \hat{y}_i = y_i \\
0 & \text{otherwise}
\end{cases}
\end{equation}

\begin{equation}
\text{Lenient Consistency} = \frac{1}{2|E|} \sum_{e \in E} \left( C_e^{\text{lit}} + C_e^{\text{fig}} \right)
\end{equation}

\textbf{Strict Consistency}: This is the most stringent evaluation. The model had to correctly identify all variations of an expression in both figurative and literal contexts to be rewarded. This setting assumes that a truly understanding model should correctly classify an idiom regardless of its context.
Let $V_e$ be the set of all variations of expression $e$.

\begin{equation}
S_e =
\begin{cases}
1 & \text{if } \forall i \in V_e, \hat{y}_i = y_i \\
0 & \text{otherwise}
\end{cases}
\end{equation}

\begin{equation}
\text{Strict Consistency} = \frac{1}{|E|} \sum_{e \in E} S_e
\end{equation}

By employing these evaluation settings, we aim to provide a comprehensive assessment of the models' capabilities in understanding and differentiating idiomatic expressions. This approach helps us determine whether the models rely on contextual understanding or memorized patterns to perform the task.

\subsection{Results}\label{sec:main_results}
\Cref{table:main_results_table} presents the results of model performances on our evaluation set. It is important to note that, GPT-4 was used to generate the sentences for DICE, so we do not consider its performance to be revealing. 

\paragraph{Degradation across evaluation levels.} As expected, performance declines from Accuracy to Lenient Consistency, and further to Strict Consistency. The results from the strictest evaluation show that, only three models—Llama 3.1, Llama 3, and GPT4o—achieve an accuracy above 40\%, with 60.36\%, 57.55\%, and 48.59\% respectively. This pattern highlights that, while the current LLMs can correctly classify individual instances, they often fail to do so consistently for both literal and figurative uses of the same idiomatic expression. Compounding this, we observe substantial standard deviations across prompts, particularly for smaller models. If the models truly relied on contextual understanding, we would expect them to perform consistently across varying levels of consistency and remain robust to prompt variations; however, the results suggest that the performance is highly inconsistent, and thus, models are not effectively leveraging context.

\paragraph{Preference towards figurative.}
The general trend for based on lenient consistency aligns with our observation on base accuracy: models show a preference for figurative interpretations when encountering an idiom, as there is a higher proportion of idioms that the models can consistently predict to be figurative across all contextual sentences than in the literal setting.

We observe the largest aggregate drop for GPT-4o which indicates that GPT-4o's high performance (evidenced by a overall accuracy of 84.33 ± 4.44 in zershot evaluations) stems from its consistency across a broad range of idioms. However, the model lacks a deep understanding of these idioms, and it is susceptible to variations. This is illustrated by an overall Lenient Consistency score of 70.32 ± 7.1, which indicates that the model can only accurately interpret a subset of idioms consistently across different texts. Llama 3.1 is the model with the least performance difference across the two subsets, indicating a more balanced understanding of both figurative and literal contexts.

\paragraph{Mixed impact of one-shot prompting.} Introducing an example improves performance for some models—most notably GPT-4o, which shows gains across all metrics and achieves the highest strict consistency (63.5\%). However, this improvement is not consistent: for many models, particularly the Flan-T5 variants, one-shot prompting offers little benefit or even degrades performance. This likely stems from a task-specific issue: instead of clarifying the task requirements, showing examples where idiomatic expressions appear literally clashes with the model’s prior expectations. Consequently, the one-shot example creates uncertainty instead of enhancing contextual understanding.

\paragraph{Summary.} While larger models like GPT-4o and Llama 3.1 (405B) show stronger overall performance, the results reveal that idiomaticity disambiguation, especially under consistency-based evaluation, remains a challenging task. The substantial drop from accuracy to strict consistency, coupled with high variation across prompts, suggests that models still rely heavily on shallow heuristics rather than robust contextual understanding.

\section{Impact of Frequency and Sentence Likelihood on Model Performance}
\label{sec:freq_likelihood}

This section examines how idiom frequency and sentence likelihood impact LLM performance. In human processing, idiom frequency and familiarity, alongside context, influence comprehension \citep{Cronk1993, Levorato_Cacciari_1992,schweigert1986comprehension,brysbaert2018word}. Similarly, in LLMs, expression frequency and sentence probability may shape idiom detection. Exploring these factors helps clarify how language-intrinsic and model-intrinsic features affect performance.

\begin{figure}[h]
\centering
\begin{adjustbox}{width=0.75\columnwidth}
\includegraphics{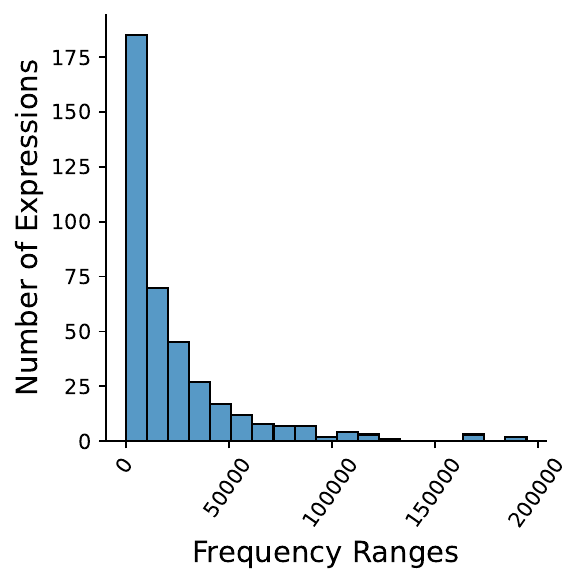}
\end{adjustbox}
  \caption{Frequency distribution of idioms in DICE below 200,000 counts.}
  \label{fig:frequency_dist}
\end{figure}

As shown in \Cref{fig:frequency_dist}, the frequency of idioms in DICE varies, with the majority of expressions occurring fewer than 200,000 times. The highest concentration of idioms falls within the 0 to 100,000 range. To focus on the most relevant portion of the dataset, we limit our analysis to this range to avoid skewed results.

\subsection{Frequency Estimation}
We use the English Web Corpus (enTenTen) \citep{jakubivcek2013tenten} to approximate the frequency of idioms in our dataset. This was for two key reasons: (1) it is parsed and tagged which allows us to query all morphological forms of the expressions using lemmas, ensuring comprehensive frequency counts, and (2) its large scale 52 billion words across diverse genres offers a robust generalization of natural language usage. While enTenTen is not identical to the pretraining datasets of LLMs, it provides a reasonable proxy for estimating expression frequency, as high-frequency terms in enTenTen are likely to appear frequently in similar web-based pretraining corpora. This analysis helps us understand whether model performance is affected by how often they may have encountered each expression during training. See \Cref{appendix:frequency_counts} for further details on how we obtain frequency counts.

\begin{figure*}
    \centering
    \scalebox{0.7}{
        \includegraphics[width=0.45\textwidth]{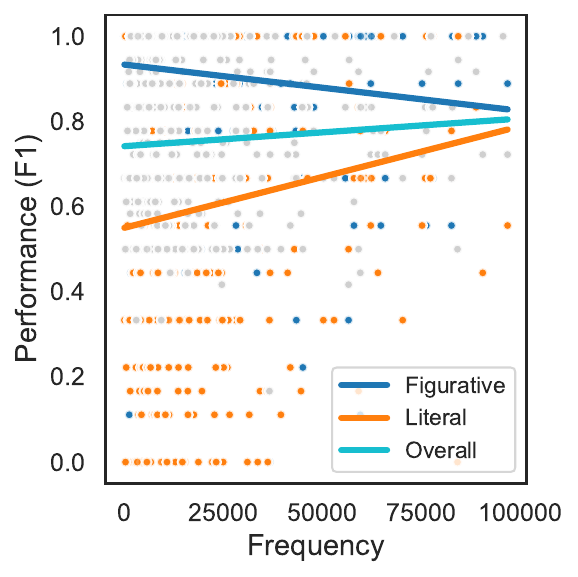}
        \includegraphics[width=0.45\textwidth]{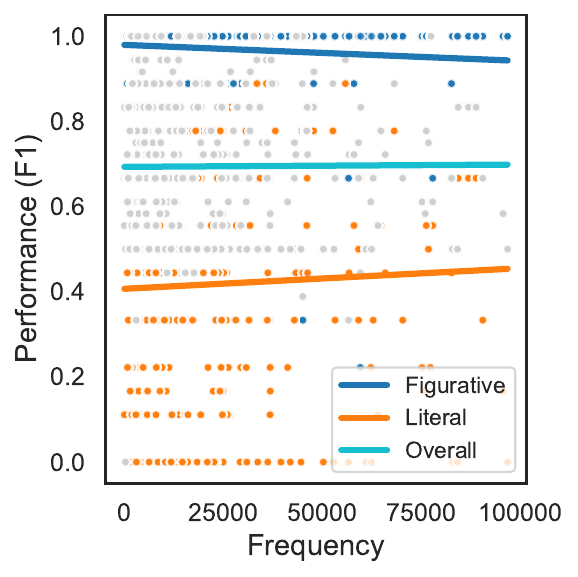}
        \includegraphics[width=0.45\textwidth]{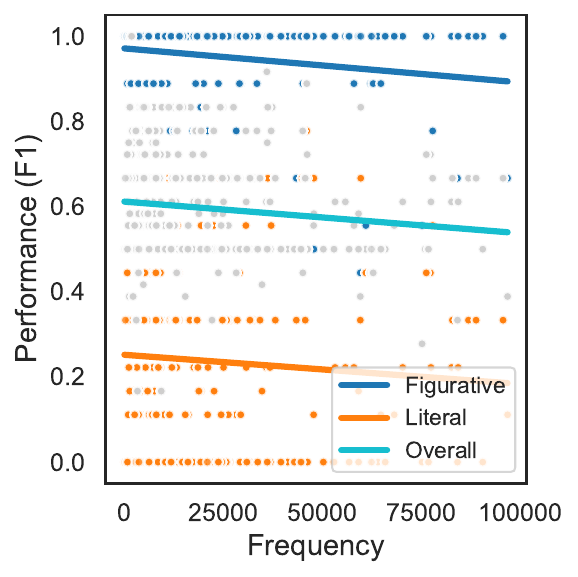}
    }
    \caption{Frequency results of GPT-3.5 Turbo, Llama 2 (70B), Flan-T5 XL (left to right).}
    \label{fig:freq_main}
\end{figure*}

\begin{figure*}
\centering
\scalebox{0.7}{
    \includegraphics[width=0.48\textwidth]{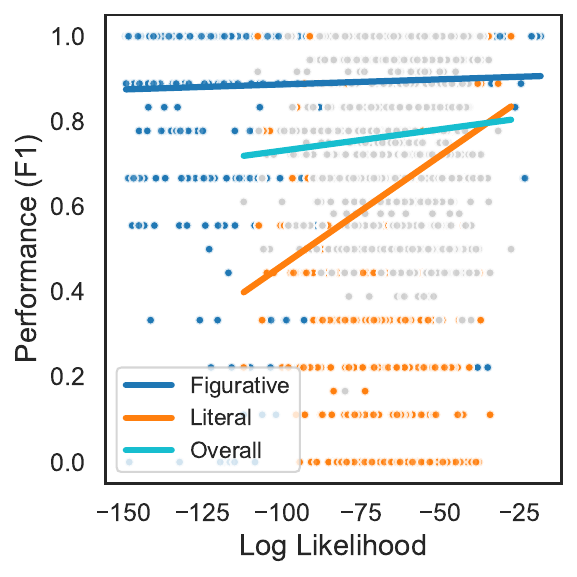}
    \includegraphics[width=0.48\textwidth]{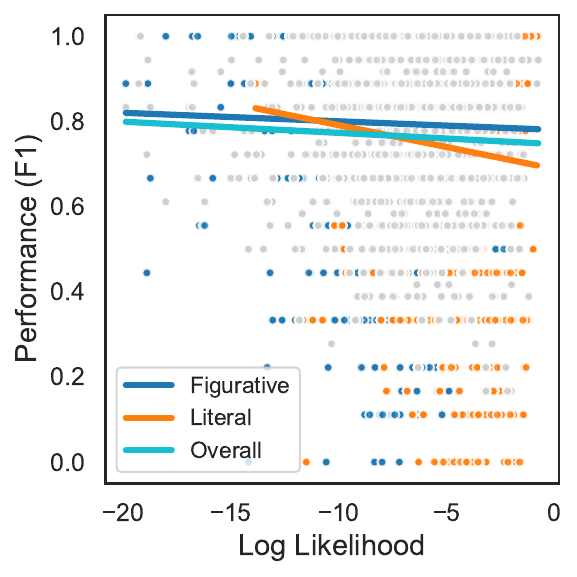}}
    \caption{Likelihood results from Llama 3 (8B) and Flan-T5 XXL (left to right).}
    \label{fig:prob_main}
   \vspace{-0.5cm}
\end{figure*}

\subsection{Likelihood Scores}
LLMs are trained to maximize the likelihood of their training data, learning to assign higher probabilities to sequences that resemble those seen during training. At inference time, these models can assign likelihood scores to input sentences, reflecting how typical or expected a sentence appears under the model’s internal distribution. Recent studies have shown that models tend to perform better, or assign higher evaluation scores, to sentences they deem more likely, regardless of the specific target task or evaluation criteria \citep{ohi-etal-2024-likelihood,mccoy-embers}.
In our setting, we use sentence-level likelihood as a proxy to test whether model performance on idiomaticity disambiguation is influenced by how probable a sentence appears to the model. Specifically, we ask whether models are more accurate on DICE when the input context has high likelihood, potentially indicating a preference for familiar or prototypical constructions over true contextual reasoning.

To analyze the relationship between sentence likelihood and model performance, we compute the likelihood of each sentence in DICE using the standard language modeling formulation. For a sentence \( y = [y^{(1)}, \dots, y^{(T)}] \), the likelihood is given by:
\[
P(y) = \prod_{t=1}^{T} P(y^{(t)} \mid y^{(<t)}),
\]
where \( y^{(<t)} \) denotes the preceding tokens. In practice, we compute the log-likelihood (negative cross-entropy loss) for each sentence and use it as a proxy for model-assigned likelihood (\Cref{sec:sent_likelihood_derivations}).

\subsection{Results}

To explore the impact of these two features on model performance, we fit a linear regression model, with idiom frequency/likelihood as the independent variable and the model's accuracy as the dependent variable. This approach allows us to estimate how these two variations influence the ability of language models to distinguish between literal and idiomatic uses. 

We present the frequency results for GPT-3.5 Turbo, Llama 2 (70B) and Flan-T5 XL in \Cref{fig:freq_main}. For the likelihood analysis, we present the results for Llama 3 (8B) and Flan-T5 XXL in \Cref{fig:prob_main}. Results for the other models can be found in \Cref{appendix:all_results_frequency_prob}. These models were chosen as they are representative of the general patterns observed across all the models evaluated.

\paragraph{Frequency is not a free lunch.} In general, we do not observe a consistent pattern that accuracy correlates with expression frequency, suggesting that while models may have encountered certain expressions frequently during pretraining, they may still struggle to properly interpret their idiomaticity in new contexts. Notably, this trend was observed in 9 out of the 13 models we analyzed. Additionally, as the frequency of an expression increases, the models tend to perform better at identifying its literal occurrences, but their accuracy in recognizing idiomatic uses declines.  One possible explanation is that for high-frequency expressions, the model may have seen both literal and idiomatic usages during pretraining. However, due to limited contextual understanding, the model may default to interpreting these expressions literally more often, regardless of the actual usage; however, this hypothesis requires further investigation. The contrasting relationship between the two settings explains why no overall correlation is observed between accuracy and frequency. Therefore, frequency does not guarantee performance.

\paragraph{Likelihood $\neq$ Understanding.} Our analysis of sentence likelihoods reveals contrasting trends across model families.  In the case of the Llama models, performance correlates positively with sentence probability. In both settings, the model perform better on sentences on which it has a higher likelihood. This is particularly the case on the literal subsets, as indicated by Llama 2 (13B) and Llama 3 (8B).
For the Flan-T5 models, we see a negative or negligible correlation between frequency/likelihood and performance. As seen in \Cref{sec:main_results}, Small and Large do not appear to have effectively utilized the context to learn meanings as successfully as the other models. The counter-intuitive pattern observed in XL and XXL models could be due to models being over-confident in their wrong prediction. This leads to situations where model would assign high probabilities to idiomatic sentences but performs poorly on the idiomaticity detection task, where surface-level fluency inflates likelihood scores without supporting deeper semantic resolution. These results show that high likelihood does not necessarily imply correct contextual understanding, especially for figurative language.

\section{Conclusion}
\label{sec:conclusion}
In this work, we contribute to idiomaticity detection in NLP with several key findings. First, we introduce DICE, a challenging dataset for context-dependent idiom detection, where distinguishing figurative from literal meanings relies heavily on understanding context. Second, we propose an evaluation framework that measures both overall accuracy and strict consistency, requiring models to correctly identify all figurative and literal instances of an expression across different contexts. Third, we show that current LLMs struggle to use context effectively, highlighting the need for models that better capture contextual nuances. 

We also investigate the effects of expression frequency and sentence likelihood. While frequency can correlate with performance in some settings, it does not guarantee accurate interpretation, highlighting that surface-level exposure is not a substitute for contextual reasoning. Similarly, while some models tend to perform better on sentences with higher likelihood, this correlation is inconsistent across models. Overall, our findings highlight the limitations of existing models in comprehending idiomatic language and highlight the need for evaluation settings, and model architectures, that emphasize deep contextual understanding over memorization or distributional familiarity.

\section{Limitations}
One of the limitations of our work is that some idiomatic expressions are noticeably more reliant on the context than others. This means that there were cases, where we could not provide a literal counterpart to the figurative interpretation. For example, the expression ``\textit{set eyes on}'' has such a dominant meaning of ``to see'', that the annotators believed to be impossible to override. In these cases, we would discard the expression. As a result, our dataset only contains a selected sample of idioms, and we acknowledge that this idea of contrastive evaluation cannot necessarily be applied to all idioms in a language.

Another of the limitations of our work is that we only consider English idioms. We would like to have extended this work to other languages, however, this relies on the existence of idiomaticity datasets in the target languages.
Moreover, the idea of making idioms literal might not be translatable to other languages, where the expression's domainant, figurative meaning cannot be overridden.

\section*{Acknowledgements}
We thank the anonymous reviewers and metareviewer for their feedback on this paper. MM is supported by the Centre for Doctoral Training in Speech and Language Technologies (SLT) and their Applications funded by UK Research and Innovation [grant number EP/S023062/1]. AV is supported by UK EPSRC grant EP/T02450X/1, The Alan Turing Institute Fellowship, the UniDive COST Action and MRC-FAPESP MR/U506734/1.  We acknowledge IT Services at The University of Sheffield for the provision of services for High Performance Computing. Additional thanks to our annotators for their hard work. A further acknowledgement to Lily Zeng, Thomas Pickard and members of the MWE Group for their helpful discussions. 

\bibliography{custom,anthology}

\appendix

\section{Examples of Expert Annotations} \label{sec:appendix_annotation}
We provid an example of expert annotations in \Cref{table:annotations}. 

\begin{table*}[hbt!]\centering
\adjustbox{max width=\textwidth}{
\begin{tabular}{|l|l|l|l|c|c|l}\toprule
Idiom &Definition of the Figurative Meaning &Sentence &Accept &Reject &Reason (if reject) \\\midrule
smoking gun &"a piece of incontrovertible evidence" &The detective found a smoking gun at the crime scene. &N &Y &Ambiguous \\
guilt trip &"to make someone feel guilty" &After breaking her mother's vase, Sarah's sister put her on a guilt trip for weeks. &N &Y &Doesn't make sense \\
turn a blind eye &"pretend not to notice" &Despite the obvious safety hazards, the supervisor chose to turn a blind eye. &N &Y &Figurative \\
down the wire &"a situation whose outcome is not decided until the very last minute" &The electrician was careful not to cut down to the wire while he was working. &N &Y &Form changed \\
set eyes on &"see" &As soon as she set eyes on the beach, she was overwhelmed by its serene beauty. &N &Y &Skip \\\midrule
blow off steam &"get rid of pent-up energy or emotion" &During the train ride, the kids were excited to see the old locomotive blow off steam. &Y & & \\
get a grip &"begin to deal with or understand" &He struggled to get a grip on the slippery glass jar of pickles. &Y & & \\
\bottomrule
\end{tabular}}
\vspace{-2mm}
\caption{\label{table:annotations}Examples of expert annotations. Definitions are taken from \citet{TheOxfordDictionaryofIdioms}. "N" and "Y" stands for "No" and "Yes", respectively.}
\end{table*}

\section{Additional Annotations of DICE}\label{appendix:addtional_summary_stats}
The analysis we focus on in this paper uses the main subsets of DICE. There are additional sentences we have collected that can be used for further analysis in the domain of idiomaticity. A summary is presented in \Cref{table:summary_stats_appendix}. We make all annotations collected publicly available.
\begin{table*}
    \centering
    \adjustbox{max width=\textwidth}{
        \begin{tabular}{lll}
        \toprule
         &Counts &Examples and Remarks \\\midrule
        All annotated sentences &2349 & This includes the aforementioned 1033 literal sentences. \\
        Unique expressions &783 &- \\
        Ambiguous sentences &165 &The panda car is a popular item in the collectible toy market. \\
        Figurative/Idiomatic sentences &465 &It was a close call when the hiker almost slipped off the cliff. \\
        Change in Form sentences &32 &She reached into the bag to find her glasses. (The idiom is "in the bag".) \\
        Doesn't make sense sentences &162 &When the children play at the park, their parents always remind them to play it safe. \\
        Grammatical Error sentences &9 &The old locomotive runs out of steam halfway up the mountain. \\
        Can't be literal sentences ("skips") &462 &The nurse cared for the critical patients day in, day out without a moment's rest. \\
        Total sentences &1295 &- \\
        \bottomrule
        \end{tabular}}
    \vspace{-2mm}
    \caption{Properties of the additional annotations that we have collected.}
    \label{table:summary_stats_appendix}
\end{table*}

\section{Sentence Generation Prompt}
\label{appendix:appendix_prompt_gen}

The prompt we used for generating the sentences is shown here. For other configurations that are not mentioned, we used the default setting.

Model: GPT-4

"role": "system", "content": "You are an expert of English"

"role": "user", "content": "Generate three sentences using the expression: '{idiom}', where the expression has a literal meaning. Each sentence must contain the expression unchanged. Format these sentences as a Python list. Don’t say anything that are not the sentences." 

The temperature used was 0.8.

\section{Participant Briefing}
\label{appendix:briefing}
Upon signing up for participation, each annotator received a 30mins training session where they were shown examples, including \Cref{table:annotations}. 
\begin{figure}[h!]
    \centering
    \scalebox{0.3}{\includegraphics{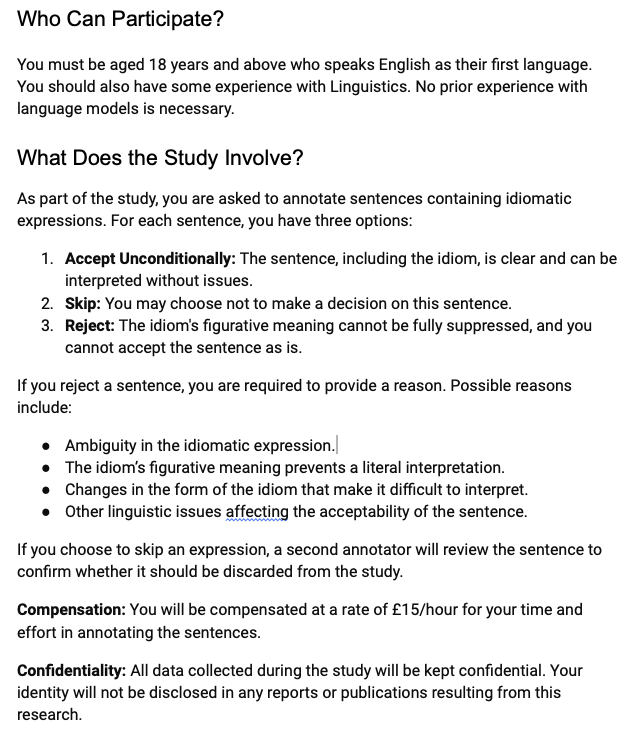}} 
    \caption{Parts of the briefing annotators received.}
\end{figure}

\section{Evaluation Implementation Details}
\label{sec:implementation_details}

\subsection{ChatGPT Versions}
\label{sec:chatgpt_versions}

We evaluate the following GPT models: GPT-4o (gpt-4o-2024-08-06) \footnote{\url{https://platform.openai.com/docs/models/gpt-4o}}, GPT-3.5-Turbo (gpt-3\_5-turbo-0125)\footnote{\url{https://platform.openai.com/docs/models/gpt-3.5-turbo}} and GPT-4 (gpt-4-0613)\footnote{\url{https://platform.openai.com/docs/models/gpt-4}}. 

\subsection{Prompting Paradigm}
\label{sec:prompting_paradigm}
We ran the FLAN-T5 models on a NVIDIA H100 GPU. We use OpenAI's API for interactions with the GPT models, HuggingFace for Flan-T5 models and Replicate \footnote{\url{https://replicate.com/}} for the Llama models. Each model was evaluated with three different prompts. All of the results we report are the average across the three prompt settings. 

\begin{table*}[!t]
    \centering
     \resizebox{\textwidth}{!}{
    \footnotesize
    \begin{tabular}{ccp{10cm}}\toprule
        Method &Prompt No. &Prompt Design \\\midrule
        \multirow{3}{*}{Zero-shot} &Prompt 1 &Is the expression '{idiom}' used figuratively or literally in the sentence: '{sentence}'. Answer 'i' for figurative, 'l' for literal. \\
        &Prompt 2 &In the sentence '{sentence}', is the expression '{idiom}' being used figuratively or literally? Respond with 'i' for figurative and 'l' for literal. \\
        &Prompt 3 &How is the expression '{idiom}' used in this context: '{sentence}'. Output 'i' if the expression holds figurative meaning, output 'l' if the expression holds literal meaning. \\
    \midrule
        Example & -- &  The expression 'play with fire' is used figuratively in the sentence: 'The war took away the unfortunate necessity, as Unionists saw it, to play with fire in the national interest, but it did not materially alter their view of themselves.' → Output: i
        The expression 'play with fire' is used literally in the sentence: 'Despite the danger, he decided to play with fire, poking the embers with a stick.' → Output: l" \\
    \midrule
        \multirow{3}{*}{Few-shot} &Prompt 1 &Example + Prompt 1\\
        &Prompt 2 & Example + Prompt 2\\
        &Prompt 3 & Example + Prompt 3\\
    \bottomrule
    \end{tabular}}
    \caption{We use three prompts for our experiments. The top panel of the table shows the prompts used in the zero-shot setting, while the bottom panel displays the few-shot prompts. For the few-shot setting, we prepend the same example (middle panel) to each of the zero-shot prompts.}\
    \label{tab:eval_prompts}
\end{table*}

\begin{figure*}[!ht]
    \includegraphics[width=.33\textwidth]{figures/appendix_figures/plots_freq_regression_v5_cameraReady/llama270bchat_.pdf}
    \includegraphics[width=.33\textwidth]{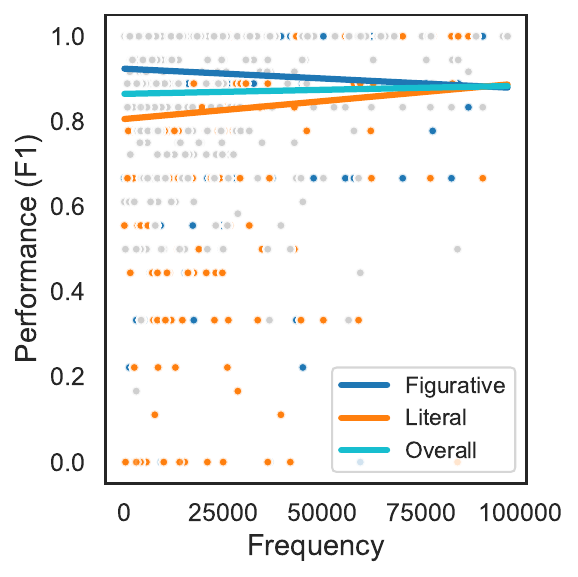}
    \includegraphics[width=.33\textwidth]{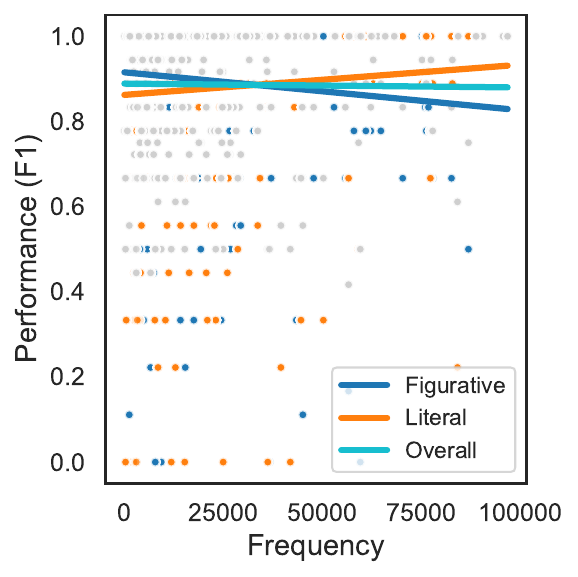}
    \caption{Left to right: Frequency analysis for Llama 2 (70B), Llama 3 (70B), and Llama 3.1 (405B)}
    \label{fig:bigger_llamas_freq_only}
   \vspace{-0.5cm}
\end{figure*}

\begin{figure*}[!hbt]
    \includegraphics[width=.33\textwidth]{figures/appendix_figures/plots_freq_regression_v5_cameraReady/gpt35turbo_.pdf}
    \includegraphics[width=.33\textwidth]{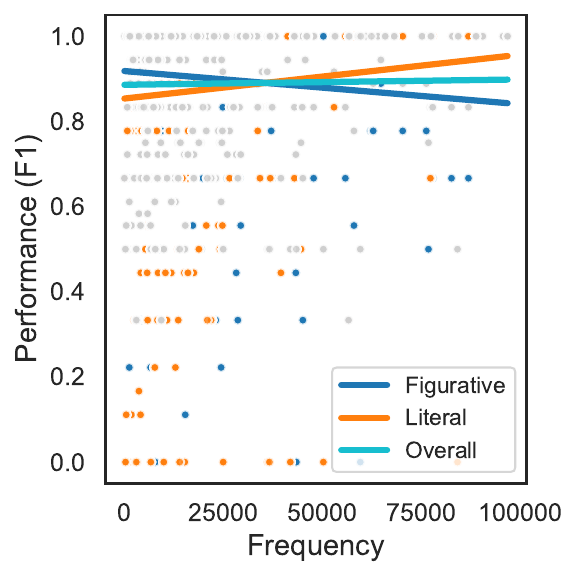}
    \includegraphics[width=.33\textwidth]{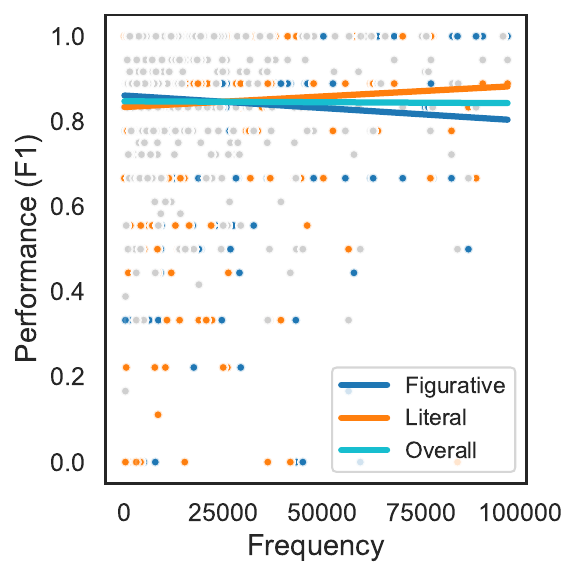}
    \caption{Left to right: Frequency analysis for GPT-3.5 Turbo, GPT-4 and GPT-4o.}
    \label{fig:gpts_freq_only}
   \vspace{-0.5cm}
\end{figure*}
\subsection{Hyper-parameters}
The hyper-parameters we used for running the evaluation are provided in \Cref{tab:hyperparams}.

\begin{table}[H]\centering
\caption{Hyper-parameters used for model evaluation, sorted by model family}\label{tab:hyperparams}
\scriptsize
\resizebox{\columnwidth}{!}{
\begin{tabular}{lrrrrrr}\toprule
&GPT Models &Flan-T5 Models &Llama 2s &Llama 3s &Llama 3.1 \\\midrule
Temperature &1 &- &0.7 &0.7 &0.6 \\
max\_tokens &Default &512 &512 &512 &512 \\
top\_p &1 &- &0.95 &0.95 &0.9 \\
\bottomrule
\end{tabular}}
\end{table}

\section{Frequency Counts}
\label{appendix:frequency_counts}
We access the enTenTen corpus using SketchEngine and employ Corpus Query Language (CQL) to find concordances that match specific lexical patterns.

For each target expression, we determine its frequency in the corpus by accounting for all lemma-based forms of the expression. We simply slot in the lemmas of the target expression directly into the following CQL query pattern. For example, for the expression “\textit{spill the beans}”, the CQL would be: {\footnotesize \texttt{[lemma=``spill''][lemma=``the''][lemma=``bean'']}}. This would capture occurrences such as “\textit{spilled the beans}”, “\textit{spilling the beans}”, and other morphological variants. We utilize NLTK \cite{bird-loper-2004-nltk} to perform lemmatisation and acquire the lemmas needed for the CQL query.

It is important to note that CQL query we use does not account for more syntactically flexible realizations, such as passive constructions (e.g., “the beans were spilled”), as these deviate from the fixed linear ordering captured by the CQL query. Consequently, the resulting frequency estimates represent a conservative measure. Nonetheless, for the purposes of correlating expression frequency with downstream model performance, we consider this approximation to be sufficiently informative.

\section{Sentence Likelihood}
\label{sec:sent_likelihood_derivations}
We derive the sentence likelihood by using the cross-entropy loss. For a sequence of tokens \( y = [y^{(1)}, y^{(2)}, \dots, y^{(T)}] \), the sentence-level cross-entropy loss \(L_{\text{sentence}}\) is defined as:
\begin{equation}
    L_{\text{sentence}} = -\sum_{t=1}^{T} \log P(y^{(t)} \mid y^{(<t)})
\end{equation}
where \(y^{<t}\) represents all tokens preceding the token at position \(t\).

Recognizing the relationship between the cross-entropy loss and the sequence probability, we observe:
\begin{align}
    L_{\text{sentence}} 
    &= -\log \left(\prod_{t=1}^{T} P(y^{(t)} \mid y^{(<t)})\right) \\
    &= -\log P(y)
\end{align}

Thus, the log likelihood of the sentence is 
\begin{equation}
    \log P(y) = -L_{\text{sentence}}
\end{equation}

We calculate the log likelihood for all sentences in DICE.

\section{Additional Results for Frequency and Likelihood Analysis}
\label{appendix:all_results_frequency_prob}
We present the results we obtained across all the models. This includes both frequency analysis per setting, and across both settings. 

Due to resource limitations we could not obtain probabilities for the larger models, which are Llama 3.1 (405B), Llama 3 (70B) and Llama 2 (70B). As a result, we only conducted the frequency analysis on these models, see \Cref{fig:bigger_llamas_freq_only}. \Cref{fig:flant5s} presents side-by-side regression plots of the frequency and likelihood analysis. Similarly, \Cref{fig:small_llamas} presents the plots for the smaller Llama models. Finally, \Cref{fig:gpts_freq_only} shows the frequency analysis plots for the GPT models.

\onecolumn

\begin{table*}[!ht]\small\centering
\setlength{\tabcolsep}{0.5pt}
\begin{tabular}{lcc}
& \texttt{Frequency Analysis} & \texttt{Likelihood Analysis} \\
\texttt{Llama 2 (7B)} &
     \raisebox{-0.6\totalheight}{\includegraphics[width=0.4\textwidth]{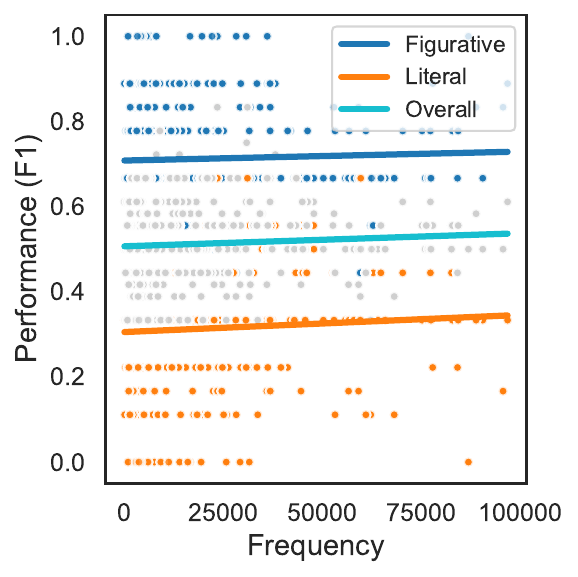}} &
    \raisebox{-0.6\totalheight}{\includegraphics[width=0.4\textwidth]{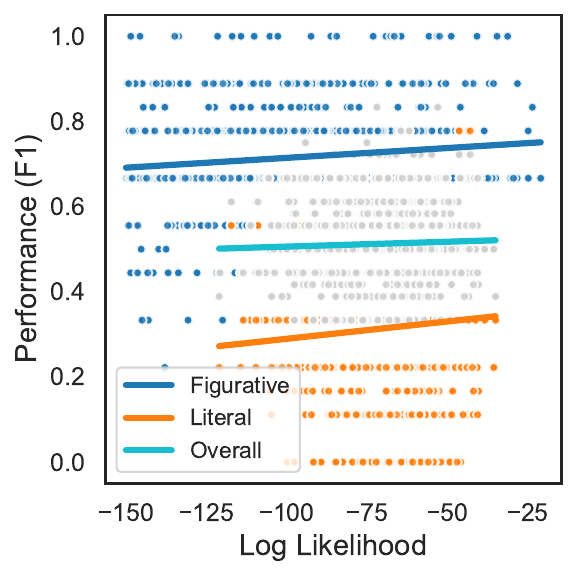}} \\
    \texttt{Llama 2 (13B)} &
     \raisebox{-0.6\totalheight}{\includegraphics[width=0.4\textwidth]{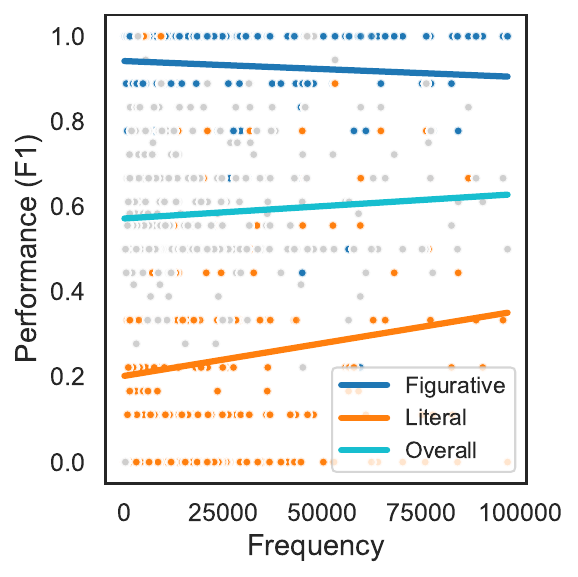}} &
    \raisebox{-0.6\totalheight}{\includegraphics[width=0.4\textwidth]{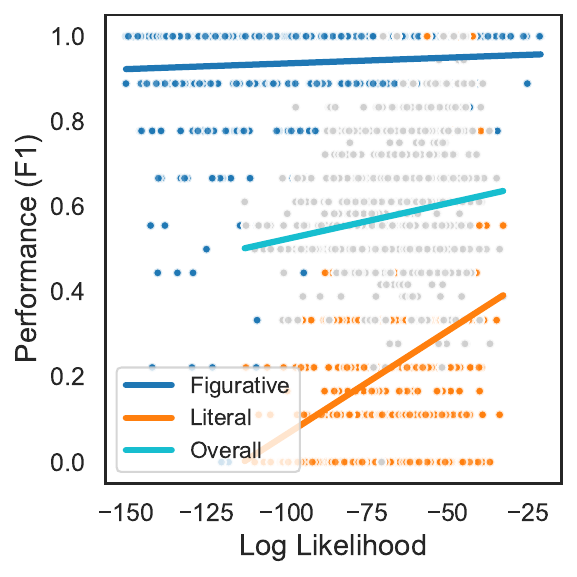}} \\
    \texttt{Llama 3 (8B)} &
     \raisebox{-0.6\totalheight}{\includegraphics[width=0.4\textwidth]{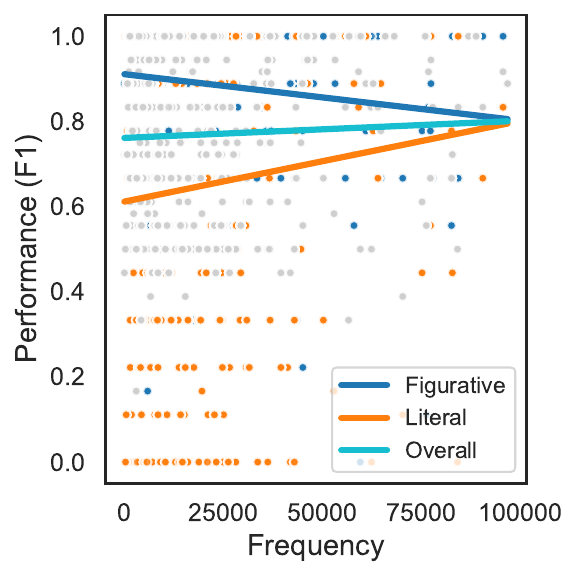}} &
    \raisebox{-0.6\totalheight}{\includegraphics[width=0.4\textwidth]{figures/appendix_figures/plots_logLikelihood_all3_selected_range_cameraReady/llama38binstruct_.pdf}} \\
\end{tabular}
\captionof{figure}{Visualisations of the frequency and likelihood. Smaller Llama models only.}
\label{fig:small_llamas}
\vspace{-0.1cm}
\end{table*}

\begin{table*}[!ht]\small\centering
\setlength{\tabcolsep}{0.5pt}
\begin{tabular}{lcc}
      & \texttt{Frequency Analysis} & \texttt{Likelihood Analysis} \\
     \texttt{Flan-T5 Small} &
     \raisebox{-0.6\totalheight}{\includegraphics[width=0.35\textwidth]{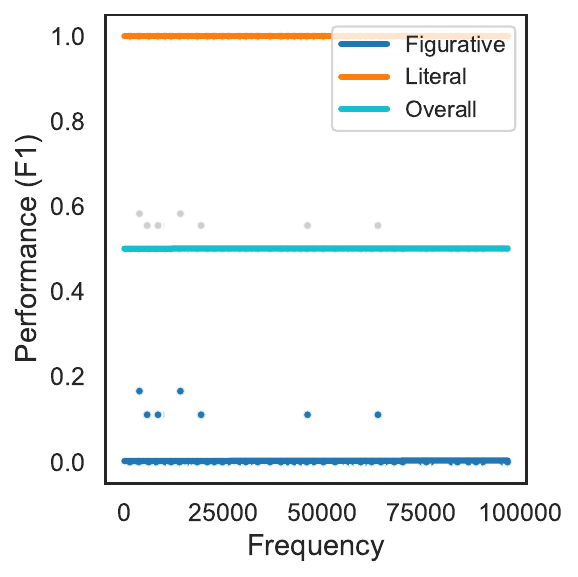}} &
    \raisebox{-0.6\totalheight}{\includegraphics[width=0.35\textwidth]{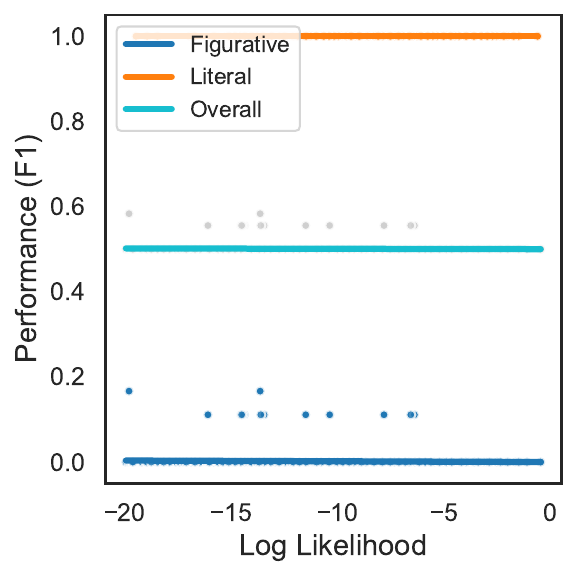}} \\
    \texttt{Flan-T5 Large} &
     \raisebox{-0.6\totalheight}{\includegraphics[width=0.35\textwidth]{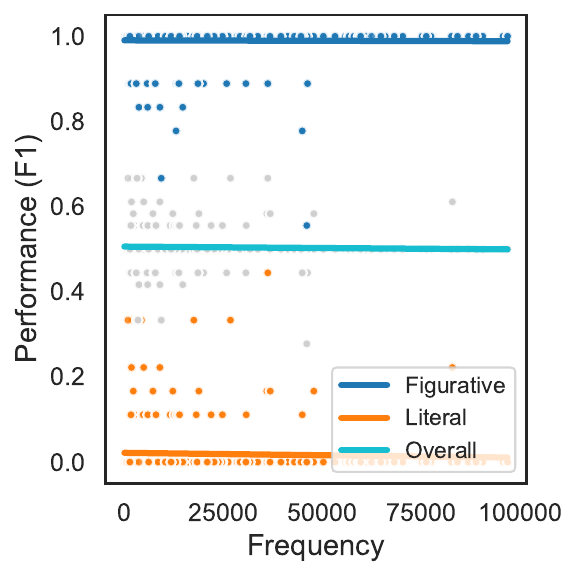}} &
    \raisebox{-0.6\totalheight}{\includegraphics[width=0.35\textwidth]{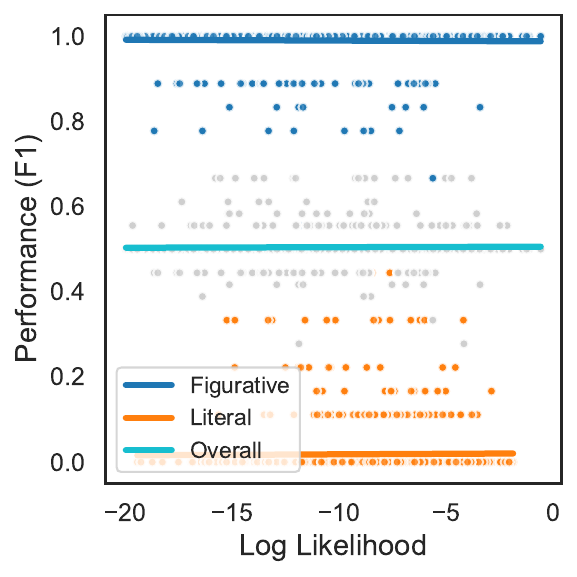}} \\
    \texttt{Flan-T5 XL} &
     \raisebox{-0.6\totalheight}{\includegraphics[width=0.35\textwidth]{figures/appendix_figures/plots_freq_regression_v5_cameraReady/flant5xl_.pdf}} &
    \raisebox{-0.6\totalheight}{\includegraphics[width=0.35\textwidth]{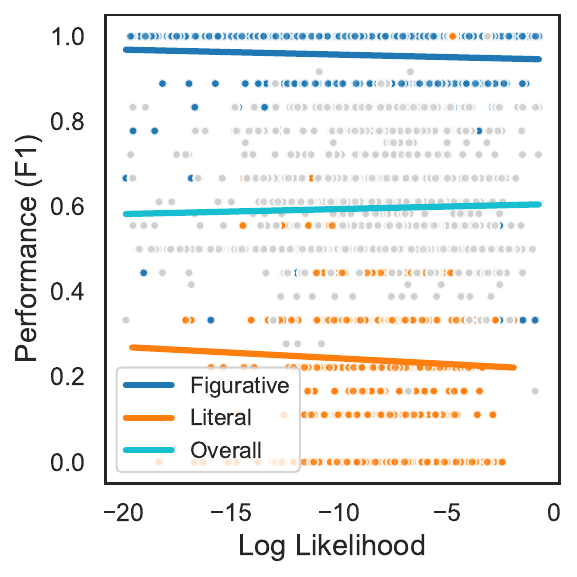}} \\
    \texttt{Flan-T5 XXL} &
     \raisebox{-0.6\totalheight}{\includegraphics[width=0.35\textwidth]{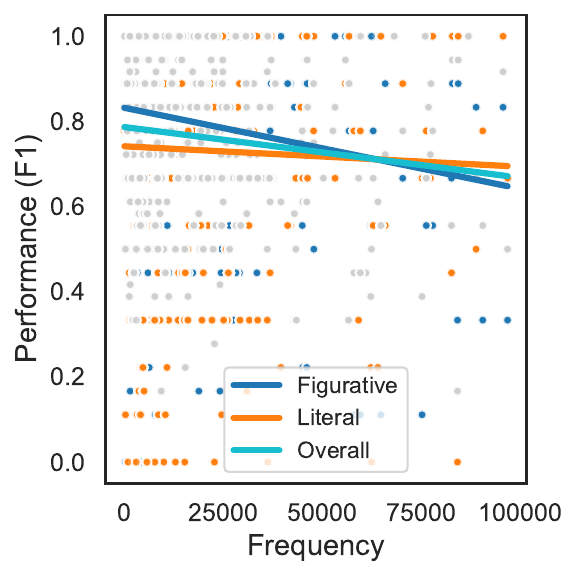}} &
    \raisebox{-0.6\totalheight}{\includegraphics[width=0.35\textwidth]{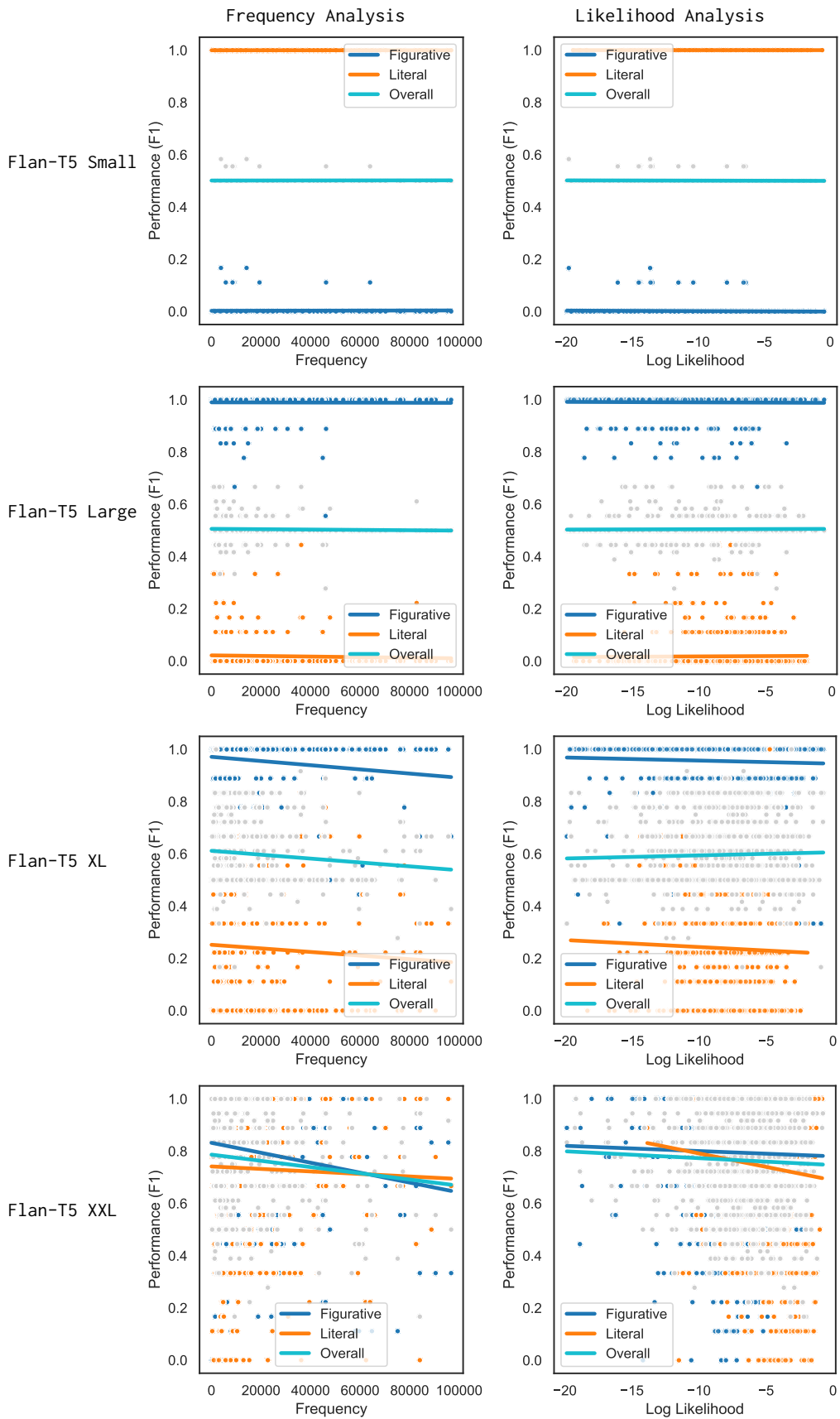}} \\
\end{tabular}
\captionof{figure}{Visualisations of the frequency and likelihood analysis. Flan-T5 models only.}
\label{fig:flant5s}
\vspace{-0.1cm}
\end{table*}

\end{document}